%% file: main.tex
\pdfoutput=1
\documentclass[
]{ceurart}

\usepackage[flushleft]{threeparttable}
\usepackage{footnote}
\makesavenoteenv{tabular}
\makesavenoteenv{table}
\usepackage{graphicx}
\usepackage{svg}
\usepackage{pdfpages}

\usepackage{caption}
\usepackage{longtable}
\begin{document}

\copyrightyear{2021}
\copyrightclause{Copyright for this paper by its authors.
  Use permitted under Creative Commons License Attribution 4.0
  International (CC BY 4.0).}

\conference{Forum for Information Retrieval Evaluation, December 13-17, 2021, India}

\title{Leveraging Transformers for Hate Speech Detection in Conversational Code-Mixed Tweets }

%


\author{Zaki Mustafa Farooqi}[%
email=zaki19048@iiitd.ac.in,
url=https://github.com/zmf0507,
]
\author{Sreyan Ghosh}[%
email=gsreyan@gmail.com,
url=https://github.com/Sreyan88,
]
\author{Rajiv Ratn Shah}[%
email=rajivratn@iiitd.ac.in,
url=http://midas.iiitd.edu.in/,
]
\address{Multimodal Digital Media Analysis Lab, Indraprastha Institute of Information Technology Delhi, India}

\input{sections/abstract}

\input{sections/keywords}
\maketitle
\input{sections/introduction}

\input{sections/related_work}
\input{sections/dataset}
\input{sections/approach}
\input{sections/results}

\input{sections/analysis}

\input{sections/conclusion}
\bibliography{main}

\end{document}

%% file: sections/abstract.tex
\begin{abstract}
In the current era of the internet, where social media platforms are easily accessible for everyone,  people often have to deal with threats, identity attacks, hate, and bullying due to their association with a cast, creed, gender, religion, or even acceptance or rejection of a notion. Existing works in hate speech detection primarily focus on individual comment classification as a sequence labelling task and often fail to consider the context of the conversation. The context of a conversation often plays a substantial role when determining the author’s intent and sentiment behind the tweet. This paper describes the system proposed by team MIDAS-IIITD for HASOC 2021 subtask 2, one of the first shared tasks focusing on detecting hate speech from Hindi-English code-mixed conversations on Twitter. We approach this problem using neural networks, leveraging the transformer's cross-lingual embeddings and further fine-tuning them for low-resource hate-speech classification in transliterated Hindi text. Our best performing system, a hard voting ensemble of Indic-BERT, XLM-RoBERTa, and Multilingual BERT, achieved a macro F1 score of 0.7253, placing us $1^{st}$ on the overall leaderboard standings.
\end{abstract}

%% file: sections/keywords.tex

%% file: sections/introduction.tex
\section{Introduction}
In today's world, hate speech is one of the major issues plaguing online social media websites. Platforms like Twitter and Gab make it easier than ever before for a person to reach a large audience quickly, which results in an increased temptation of users for inappropriate behavior such as hate speech, causing potential damage to the social system and thus possessing major threats which has already led to different types of crimes \citep{williams2020hate}. Human moderators manually detecting hate speech online have been reported to go through trauma and mental issues. This phenomenon necessitates automated hate speech detection as a crucial task.

A majority of the work on hate speech classification is constrained to the $English$ language. The inability of mono-lingual hate speech classifiers to detect the semantic cues in code-mixed languages necessitates an efficient classifier that can detect offensive content
automatically from code-mixed languages. Hinglish (formed of the words spoken in Hindi language but written in Roman script
instead of the Devanagari script) extends its grammatical setup from native Hindi, accompanied by many slurs, slang, and phonetic variations due to regional influence. Randomized spelling variations and multiple possible interpretations of Hinglish words in different contextual situations make it extremely difficult to deal with for automated classification. Another challenge worth considering in dealing with Hinglish is the demographic
divide between Hinglish users relative to total active users globally. This poses a severe limitation as the tweet data in Hinglish language is a small fraction of the large pool of tweets generated, necessitating the use of selective methods
to process such tweets in an automated fashion.

\begin{figure}[htbp]
  \centering
    \includegraphics[width=0.8 \textwidth]{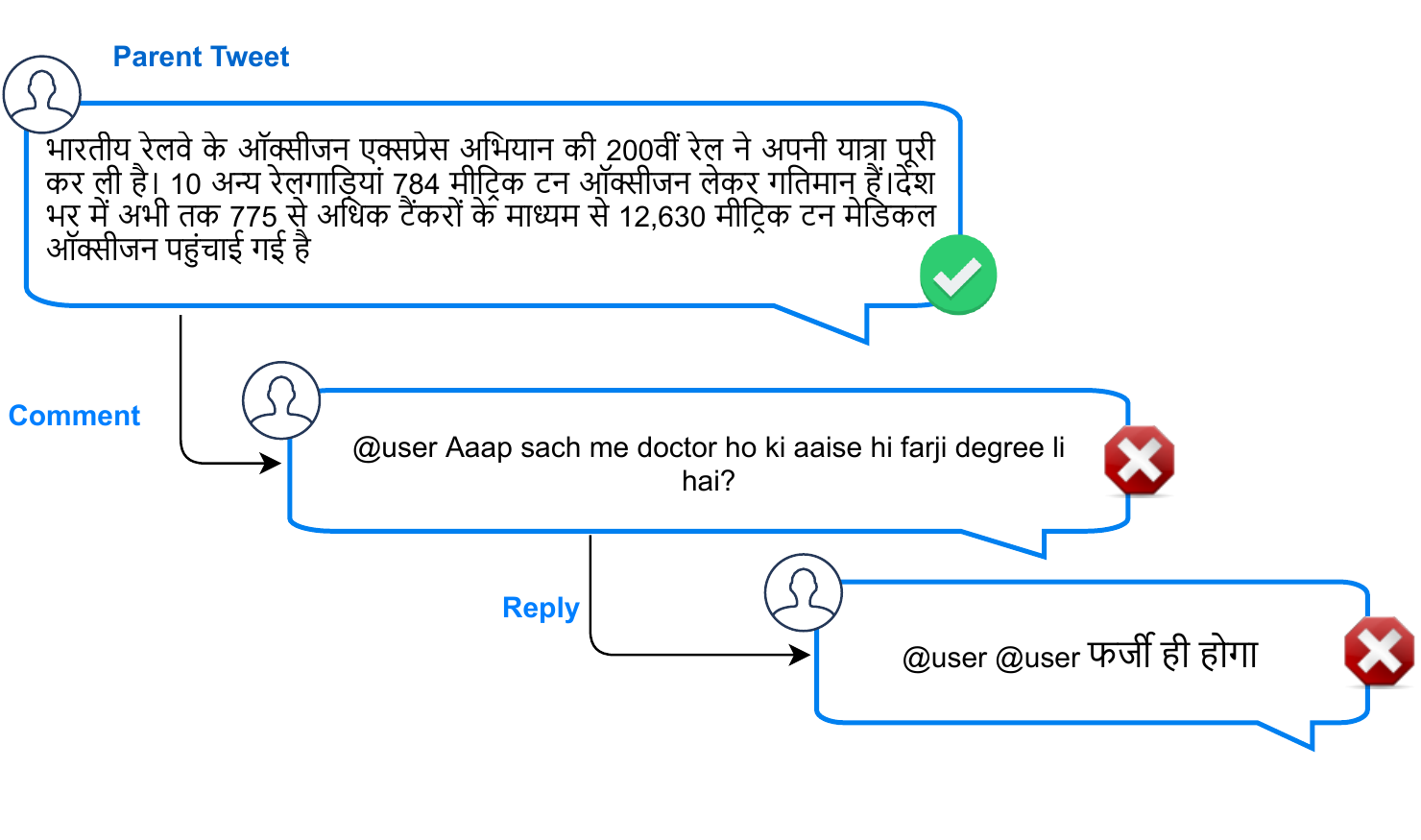}
  \caption{An example of conversation from the dataset where the parent tweet is not hateful but the comment and reply are expressing implicit hate towards the user who posted parent tweet. The post explains the amount of oxygen that has been supplied across the whole country. The hateful comment says ``Are you a genuine doctor or have you just acquired a fake degree?'' while the hateful reply gives an affirmative response by saying ``it must be fake''. 
  The green tick represents ``NOT'' comments while the red cross represents ``HOF'' or hateful/offensive comments.}
   \label{fig:hate-example}
\end{figure}

The context of the conversation plays a very crucial role in hate speech identification. We acknowledge the fact that a comment categorized as hate speech may not always contain the subject of hate on its own. As shown in figure \ref{fig:hate-example}, agreement or disagreement with a previous comment or the overall ideology in the chain of the conversation might also induce hate towards a particular target group. In addition to this, systems that can efficiently utilize the entire context of the comment chain with a holistic understanding of the entire discussion may also help in detecting ``trigger comments'' i.e., non-toxic comments in online discussions which lead to toxic replies and implicitly help in mitigating bias in hate speech identification which remains a long-standing problem in this domain \citep{dixon2018measuring,borkan2019nuanced}.

Hate Speech and Offensive Content Identification in English and Indo-Aryan Languages (HASOC) 2021 \cite{hasoc2021_overview} proposes two subtasks where subtask 1  aims towards identifying and discriminating hate and profane tweets in English, Hindi, and Marathi, and subtask 2 aims towards detecting hate tweets in conversations primarily in Hindi-English code-mixed texts. This paper illustrates our key contribution to subtask 2 \cite{hasoc2021_subtask2} of \textbf{HASOC 2021} \cite{hasoc2021_overview}. We present our system based on ensemble of transformers to detect hate speech in code-mixed Hindi-English conversations, which helps us achieve the first place in the HASOC Subtask 2 final leaderboard standings.

%% file: sections/related_work.tex
\section{Related Work}

Hate speech detection is a challenging task with literature including techniques such as dictionary-based \citep{4618794}, distributional semantics \citep{10.1145/2740908.2742760} and recent literature exploring the power of neural network architectures for the same \citep{badjatiya2017deep}. However, a majority of the work done on hate speech detection is constrained to the English language \citep{founta2018large, carta2019supervised,saeed2018overlapping,vaidya2020empirical,tran2020habertor,hitkul2019maybe,aggarwal2020trawling, ghosh2021speech} with very limited work on other foreign languages \citep{kamal2021hostility,leite2020toxic,saroj2020indian,basile-etal-2019-semeval,ghosh-chowdhury-etal-2019-arhnet} and code-switched text \citep{mathur2018detecting,kapoor2019mind,mathur2018did,chopra2020hindi,kamble2018hate}. Although Hinglish has been a major contributor to hate speech online, this area has seen very little work with recent work exploring transformers \citep{ranasinghe2020wlvrit} and author profiling using graph neural networks \citep{chopra2020hindi}. Works like \citep{mathur2018detecting} uses a Convolutional Neural Network (CNN) architecture togther with Glove embeddings and transfer learning, in one of the first attempts to detect hate-speech online in the \emph{Hinglish} language. Similar to our work, \citep{yasaswini-etal-2021-iiitt} explored XLM-RoBERTa and achieved competitive results on the task of detecting hate-speech in Dravidian languages. In HAOSC 2020 shared task, \citep{baruah2021iiitgadbuhasocdravidiancodemixfire2020} also used XLM-RoBERTa to achieve the third rank on the overall task of detecting offensive content in code-mixed Dravidian text. However, we acknowledge the fact that XLM-RoBERTa was pre-trained on Hindi only text and our work differs from theirs in which we also transliterate words in a different code to Hindi to solve the problem of hate-speech detection.

Hate speech detection has branched into several sub-tasks like toxic span extraction \citep{pav2020semeval,ghosh2021cisco}, rationale identification \citep{mathew2020hatexplain} and hate target identification \citep{basile-etal-2019-semeval}. Though recent advancement in the field of NLP has pushed the limits of hate speech identification, like transformers \citep{chopra2020hindi} and graph neural networks \citep{mishra2019abusive,chopra2020hindi,das2021brutus} with people attempting to induce external knowledge leveraging author profiling \citep{chopra2020hindi} or ideology \citep{qian2018hierarchical} but using context of the conversation is still a challenge with very little work exploring this problem. Context of the conversation plays a huge role in hate speech identification with recent literature exploring both the structure and effect of context \citep{pavlopoulos2020toxicity,saveski2021structure} for the same. One interesting and related direction of work described in \citep{qian2019benchmark} relates to building systems that generate text which acts as hate speech interventions in online discussion. Context can both help in detecting trigger comments \citep{10.1145/3366423.3380074} and implicitly handle bias in hate speech identification which remains a long-standing problem in this domain \citep{dixon2018measuring,borkan2019nuanced}. To the best of our knowledge, all these works are constrained to the English language with very little work on code-mixed Hindi-English and Hinglish text, considering the context of the conversation.

%% file: sections/dataset.tex
\section{Dataset}

The dataset provided for this task has code-switched Hindi-English as well as Hinglish conversation chains taken from twitter as shown in Fig 1. This is a binary classification dataset having two classes HOF and NOT.      
HOF denotes the tweet, comment, or reply which contains hate, offensive, and profane content in itself or is supporting hate, whereas NOT denotes the tweet, comment, or reply which does not contain any hate speech, profane, or offensive content. More details about the dataset provided to us can be found in table \ref{tab:orig-dataset}. We have also provided \emph{Avg. Comments} which tells us about the average number of comments to a \emph{Parent Tweet} in the dataset since all \emph{Parent Tweets} had at least one \emph{comment}. We do not provide the average number of replies for comments since all comments do not have replies.

\input{sections/dataset-stats-table}

\begin{table}[h]
\centering
\caption{Train-Validation-Test Distribution}
\label{tab:proc-dataset}
\begin{tabular}{@{}cccc@{}}
\toprule
\textbf{Data} & \textbf{Total Conversations} & \textbf{HOF(Hateful/Offensive)} & \textbf{NOT (neither Hateful nor Offensive)} \\ \midrule
Train         & 4592                         & 2273                            & 2319                                         \\
Val           & 1148                         & 568                             & 580                                          \\
Test          & 1348                         & 695                             & 653                                          \\ \bottomrule
\end{tabular}
 \begin{tablenotes}
      \item Note : \textbf{Val} refers to Validation data .
    \end{tablenotes}
\end{table}

For feeding data into our model, we flatten the given conversations into individual parent-comment-reply unique conversation chains. As mentioned earlier, all parent tweets have atleast one comment, but all comments do not have replies, so,  a training instance might end with a comment. For each instance, the final label assigned to the instance was the label of the final comment in the conversation chain. Table \ref{tab:proc-dataset} describes the dataset distribution with the validation split used in our experiments in section \ref{eval-valid} .

%% file: sections/dataset-stats-table.tex
\begin{table}[h]
\centering
\caption{Original Dataset Statistics}
\label{tab:orig-dataset}
\resizebox{\textwidth}{!}{%
\begin{tabular}{@{}cccccccc@{}}
\toprule
\textbf{Data} & \textbf{Total  Conversations} & \textbf{HOF} & \textbf{NOT} & \textbf{Parent Tweets} & \textbf{Total Comments} & \textbf{Total Replies} & \textbf{Avg. Comments} \\ \midrule
Train         & 5740                                & 2841         & 2899         & 82                     & 3778                    & 1880                   & 46                     \\
Test          & 1348                                & 695          & 653          & 16                     & 849                     & 483                    & 53                     \\ \bottomrule
\end{tabular}%
}
 \begin{tablenotes}
      \item Note : \textbf{Val} refers to Validation data .
    \end{tablenotes}
\end{table}

%% file: sections/approach.tex
\section{Methodology}
Our methodology primarily involves fine-tuning the transformer models pre-trained on a massive multilingual corpus. The following sections further describe the end-to-end approach used in our experiments. 
\subsection{Data Pre-processing}
We first start by concatenating the tweet and its comments and replies, if they are present, to form the final text sequence. Our intuition behind this process is that this concatenation will help the model understand the context better, especially in those cases where the comment or reply may not be hateful but shows support for the hateful parent tweet. 
While concatenating the tweets, we insert a new separator token ``[SENSEP]'' between the tweet , comment, and the reply, to differentiate between one tweet and another.
Post concatenation, we perform data cleaning by removing hashtags, emojis, URLs and mentions from the tweets. However, we do not remove punctuation and numbers to preserve the syntactic and semantic coherence of the tweets. 

Although the data is code-mixed, having both Hindi and English text, there were several instances where Hindi text is present in Roman script. Since our models are pretrained on a code-mixed corpus, it is essential to deal with Hindi text in the Roman script to make the whole dataset consistent for training purposes. Therefore, we perform transliteration using AI4Bharat library\footnote{ \url{https://pypi.org/project/ai4bharat-transliteration/}} to convert the Hindi text in Roman script to Devanagari script. 

\begin{figure}[htbp]
  \centering
  \includegraphics[width= \textwidth]{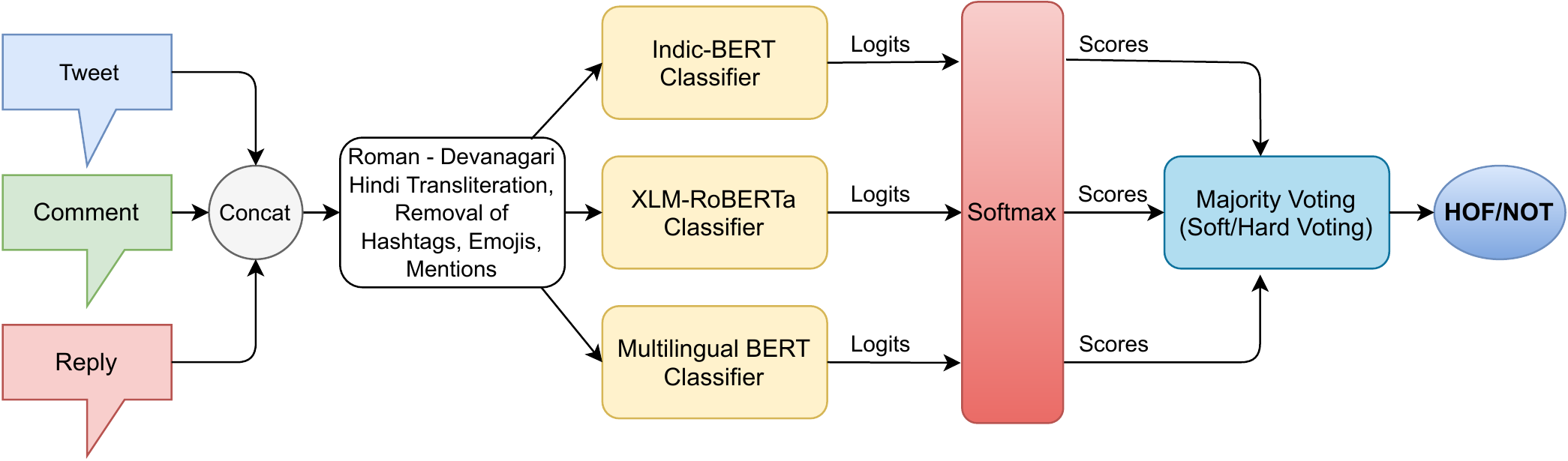}
  \caption{Overview of Ensemble Model. 'Concat' refers to concatenation operation. 'Scores' refer to normalized logits/ probability scores .}
  \label{fig:model_image}
\end{figure}



\subsection{Baseline Approach : Fine-tuning Transformer Models}
\label{models} 
We perform experiments with three different types of transformer models, which are described below.

\begin{itemize}
    \item \textbf{Indic-BERT} \cite{indic-bert}  is an ALBERT model pre-trained on a massive multilingual corpus having 12 Indic languages such as Assamese, Bengali, English, Gujarati, Hindi, Kannada, Malayalam, Marathi, Oriya, Punjabi, Tamil, Telugu \cite{indic-bert} . The multilingual corpus has about 9 billion tokens. 
    
    \item \textbf{Multilingual BERT} (mBERT) is a BERT \cite{devlin2019bert} model pretrained on Wikipedia data having over 100 languages with a masked language modeling objective. 
    
    \item \textbf{XLM-Roberta} (XLM-R)  \cite{xlm-roberta} is a  transformer-based masked language model pretrained on Common Crawl data having about 100 languages. It was proposed by Facebook and happened to be one of the best-performing transformer models for multilingual tasks.

\end{itemize}

On top of these pre-trained transformer models, we add a dropout followed by a fully connected layer of size two which takes in the transformer's CLS token's representation of size 768. The fully connected layer returns logits for the two classes, i.e., HOF and NOT, which are then passed to a softmax layer to predict the class of the input text. This is explained as a part of the ensemble model in figure \ref{fig:model_image}
 .
\subsection{Our Approach : Models Ensembling}
We perform model ensembling on the top of our three fine-tuned models as explained in the section \ref{models}. Since we had three different transformer models fine-tuned for our task, we decided to combine the output of the three models using the majority voting system, i.e., Hard Voting and Soft Voting. We denote our two ensemble models as \textbf{Hard Voting Ensemble} and \textbf{Soft Voting Ensemble}. \textbf{Hard Voting Ensemble} model takes in the class predictions from each of the fine-tuned transformer models and selects the class with maximum votes. Similarly, \textbf{Soft Voting Ensemble} model takes in the class probabilities from each of the fine-tuned transformer models and sum the same class probabilities and selects the class having higher probabilities sum. Figure \ref{fig:model_image} shows the end-to-end model pipeline used in our experiment.

%% file: sections/results.tex
\section{Experiments and Results}
\subsection{Experimental Setup}
For fine-tuning our transformer models, we set a learning rate of 2e-5 for all our experiments with AdamW \cite{adamw} optimizer and a linear learning rate scheduler. We train our models with a batch size of 8. Our experiments use the Hugging-Face Transformers library \cite{hugging-face} for fine-tuning all the pre-trained transformer models.

\subsection{Evaluation with Validation data}
\label{eval-valid}

In order to evaluate our models, we split the train dataset in an 80:20 ratio where 80\% of the data is the new train data, and the rest of 20\% becomes the validation data as shown in table \ref{tab:proc-dataset}. The split is done in random order while maintaining the same class ratio in both train and validation data.
We train our models for ten epochs and keep track of the validation loss at each epoch. For reporting results on validation data, We use the model checkpoint corresponding to the epoch with minimum validation loss. At the same time, we make a note of that epoch for which we later train our model on the whole train dataset in \ref{eval-test}.

\input{sections/val-table}

From Table \ref{tab:val-results} , we can see that our model ensembling approaches \textbf{Soft Voting Ensemble} and \textbf{Hard Voting Ensemble} outperform rest of the baseline transformer models with Macro F1 score of \textbf{.7682} and \textbf{.7621} respectively.

\input{sections/test-table}

\subsection{Evaluation with Test data}
\label{eval-test}
Following the results obtained in section \ref{eval-valid} , we train our models on the whole train dataset for the best number of epochs identified through the minimum validation loss in section \ref{eval-valid} . These epochs vary from 3 to 6 for each of the three transformer models discussed in \ref{models}.
We submit the test results for three of our models as marked in table \ref{tab:test-tab} which reports the results obtained on the test data. It can be inferred that the test results follow the same trend as was with the validation data. However, \textbf{Hard Voting Ensemble} with Macro F1 of \textbf{0.7253} is slightly better than \textbf{Soft Voting Ensemble} with Macro F1 score of \textbf{0.7223}. However, the overall performance of the model ensembling technique is better than the Indic-BERT, XLM-RoBERTa, and Multilingual BERT by a significant margin where the highest possible Macro F1 score for baseline models is \textbf{0.7031}.

%% file: sections/val-table.tex
\begin{table}[h]
\centering
\caption{Results obtained on Validation split }
\label{tab:val-results}
\resizebox{\textwidth}{!}{%
\begin{tabular}{ccccc}
\hline
\textbf{Model} &
  \textbf{F1} &
  \textbf{Precision} &
  \textbf{Recall} &
  \textbf{Accuracy(\%)} \\ \hline

Indic-BERT                          & 0.7150           & 0.7159   & 0.7154   & 71.51 \\
Multilingual BERT                   & 0.7438         & 0.7438 & 0.7439 & 74.39 \\
XLM-RoBERTa                         & 0.7262           & 0.7277   & 0.7268   & 72.64 \\
\rowcolor[HTML]{EFEFEF} 
Soft Voting Ensemble (Our approach) & \textbf{0.7682}  & 0.7687   & 0.7684   & 76.82 \\
\rowcolor[HTML]{EFEFEF} 
Hard Voting Ensemble (Our approach) & \textbf{0.7621} & 0.7628   & 0.7624   & 76.21 \\ \hline
\end{tabular}%
}
 \begin{tablenotes}
      \item Note : F1, Precision and Recall are Macro Scores
 \end{tablenotes}
\end{table}

%% file: sections/test-table.tex
\begin{table}[h]
\centering
\caption{Results obtained on Test Data}
\label{tab:test-tab}
\resizebox{\textwidth}{!}{%
\begin{tabular}{lccccc}
\hline
\textbf{Submission} & \textbf{Model}                   & \textbf{F1}     & \textbf{Precision} & \textbf{Recall} & \textbf{Accuracy(\%)} \\ \hline
NA                  & Indic-BERT                          & 0.6811          & 0.6881             & 0.6821          & 68.47                 \\
NA                  & Multilingual BERT                   & 0.7031          & 0.7031             & 0.7033          & 70.33                 \\
submit-1            & XLM-RoBERTa                         & 0.6970          & 0.6970             & 0.6970          & 69.73                 \\
\rowcolor[HTML]{EFEFEF} 
submit-2            & Soft Voting Ensemble (Our approach) & \textbf{0.7223} & 0.7236             & 0.7222          & 72.32                 \\
\rowcolor[HTML]{EFEFEF} 
submit-3            & Hard Voting Ensemble (Our approach) & \textbf{0.7253} & 0.7267             & 0.7251          & 72.62                 \\ \hline
\end{tabular}%
}
 \begin{tablenotes}
      \item Note : F1, Precision and Recall are Macro Scores
 \end{tablenotes}
\end{table}

%% file: sections/analysis.tex
\section{Analysis}

\begin{figure}[htbp]
\centering
\scalebox{0.6}{
\includegraphics[width=\textwidth]{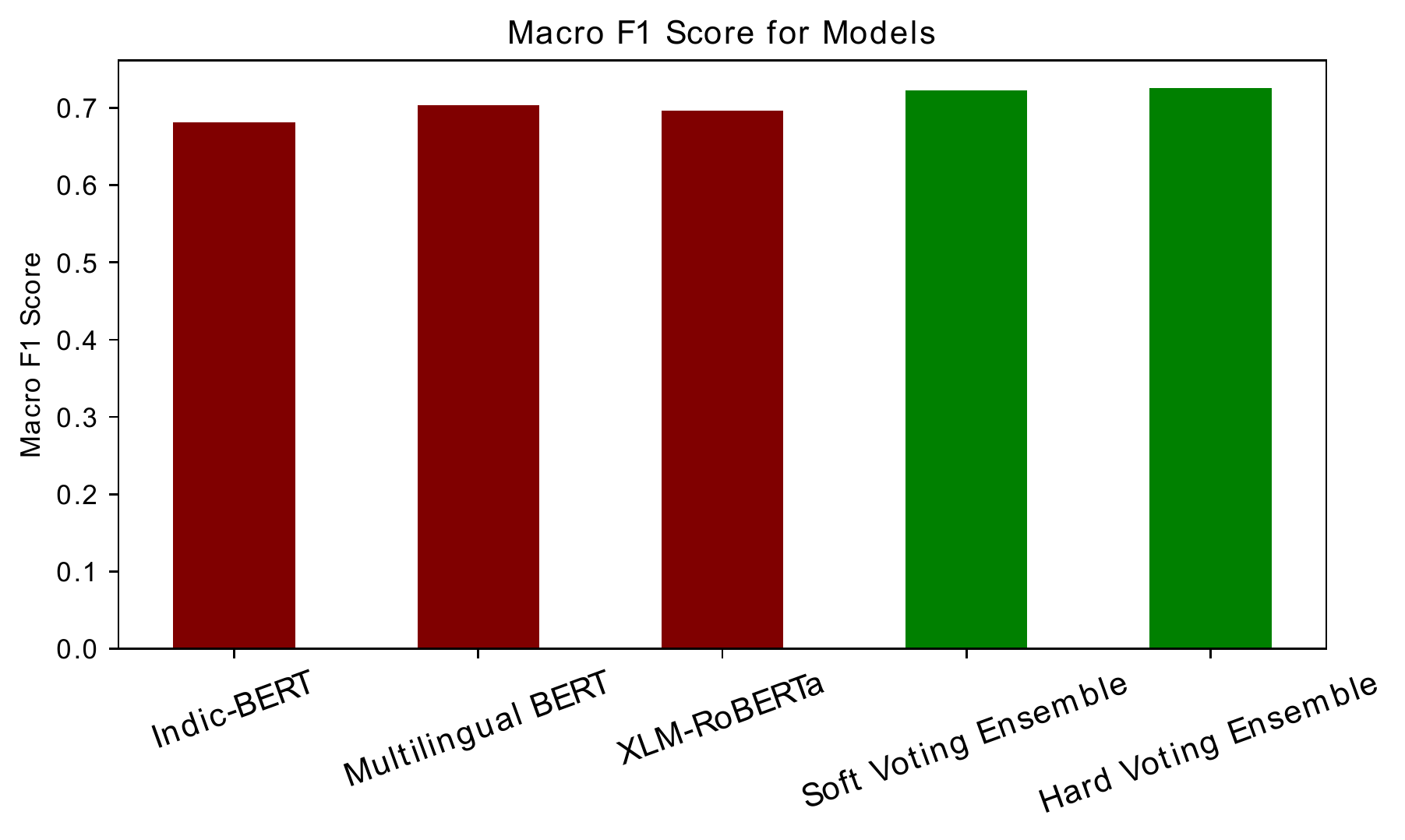}
}
\caption{Macro F1 Scores on Test Data}
\label{fig:f1_bar}
\end{figure}

Figure \ref{fig:f1_bar} compares the results in table \ref{tab:test-tab}. It can be noted that the Indic-BERT model has the lowest F1 score among all the baseline and ensemble models. \textbf{Soft Voting Ensemble} and \textbf{Hard Voting Ensemble} models yield better results than all the baseline transformer models. However, merely having a better F1 score is not enough since it is crucial to understand where our approach fails and where it performs better than the baselines. 

We start our error analysis with table \ref{tab:misclassify_table} where we have shown the total number of misclassified samples from each class. In addition to it, it has the percentage of total samples misclassified from each class, which helps us develop a better understanding of the direction where models are making more mistakes. Figure \ref{fig:cm_models} shows the detailed results of the performance of each model for samples of both classes. A closer look reveals that Multilingual BERT and XLM-RoBERTa misclassify almost equal number of samples from both classes HOF and NOT where the percentage ranges from 29.35\% to 31.24\%. 
However, the case is quite different for Indic-BERT, where the misclassification rate is very high for the NOT class, which is 40.27\% compared to the misclassification rate of 23.30\% for the HOF class. This could also be because Indic-BERT is trained on 12 Indian languages and has a less diverse pre-training corpus than the other two transformer models. 

\input{sections/misclassify_table}
\input{sections/confusion_matrix}

As far as our ensemble models are concerned, we observe that the misclassification rate is very balanced for both classes. To compare it with the baseline transformer models, we can see that the ensemble models have the lowest misclassification rate for HOF class ranging between 23\% to 24\%, which was also the case with Indic-BERT, and similarly,  the misclassification rate for NOT class is also close to the lowest among the baseline models. 
This suggests that model ensembling minimized the misclassification rate for both classes, and we can further conclude that a single model's mistake is likely to be corrected by the other two models in an ensemble model. However, the ensemble models still make 7-8\% more mistakes in identifying NOT class compared to HOF class.

%% file: sections/misclassify_table.tex
\begin{table}[h]
\centering
\caption{Total Number and Percentage of Misclassified samples for HOF and NOT classes }
\label{tab:misclassify_table}
\resizebox{\textwidth}{!}{%
\begin{tabular}{@{}ccccc@{}}
\toprule
\textbf{Model} & \textbf{Misclassified HOF} & \textbf{\% MR (HOF)} & \textbf{Misclassified NOT} & \textbf{\% MR (NOT)} \\ \midrule
Indic-BERT           & 162 & 23.30 & 263 & 40.27 \\
Multilingual BERT    & 207 & 29.78 & 193 & 29.55 \\
XLM-RoBERTa          & 204 & 29.35 & 204 & 31.24 \\
Soft Voting Ensemble & 168 & 24.17 & 205 & 31.39 \\
Hard Voting Ensemble & 165 & 23.74 & 204 & 31.24 \\ \bottomrule
\end{tabular}%
}

 \begin{tablenotes}
      \item Note : \% MR (HOF) and \% MR (NOT) refer to percent misclassification rate for HOF and NOT classes respectively. \% MR (class) is the percentage of total misclassified samples of the class out of the total samples of the class. These are calculated with total HOF samples count of 695 and total NOT samples count of 653 in test data.
\end{tablenotes}
\end{table}

%% file: sections/confusion_matrix.tex
\begin{figure}[hbt!]
\centering
\includegraphics[width=.18\textwidth]{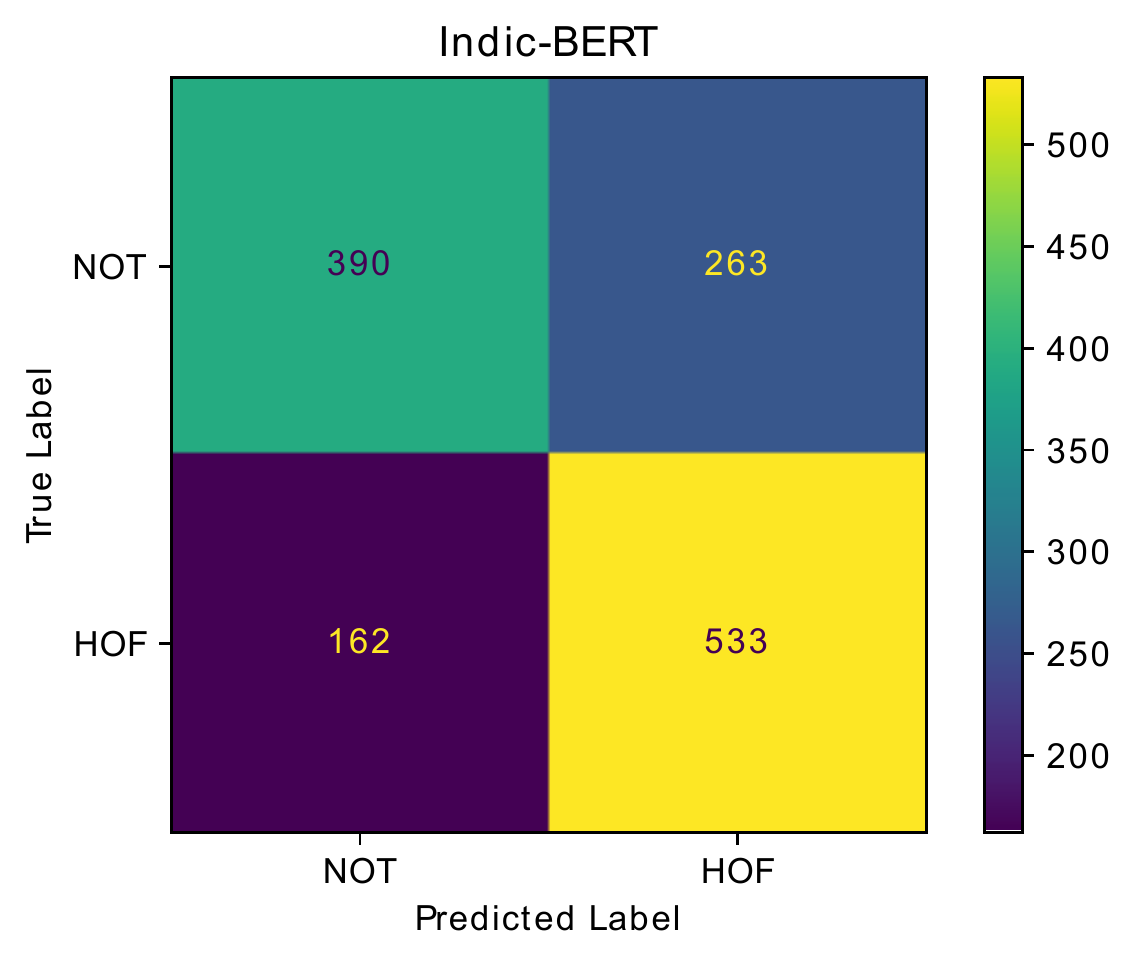}\hfill
\includegraphics[width=.18\textwidth]{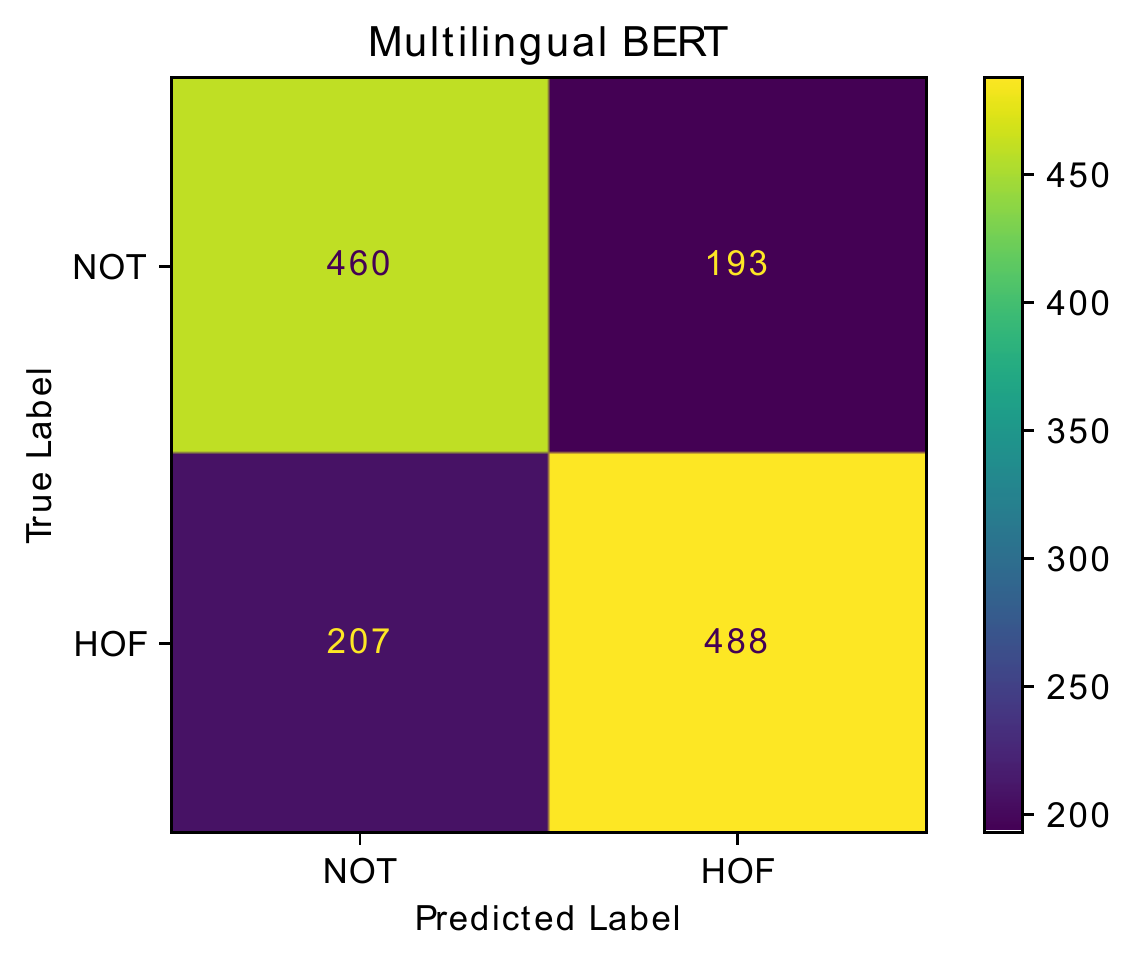}\hfill
\includegraphics[width=.18\textwidth]{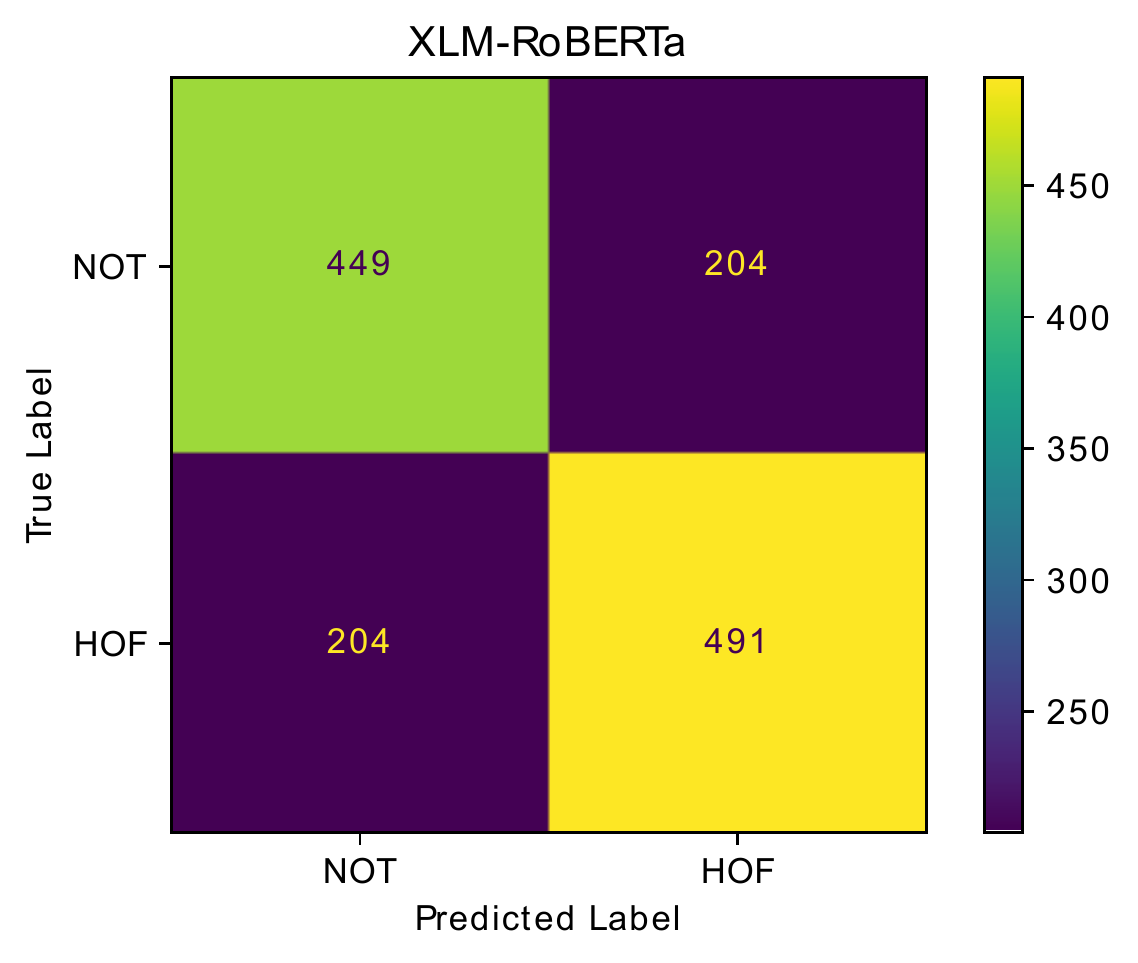}\hfill
\includegraphics[width=.18\textwidth]{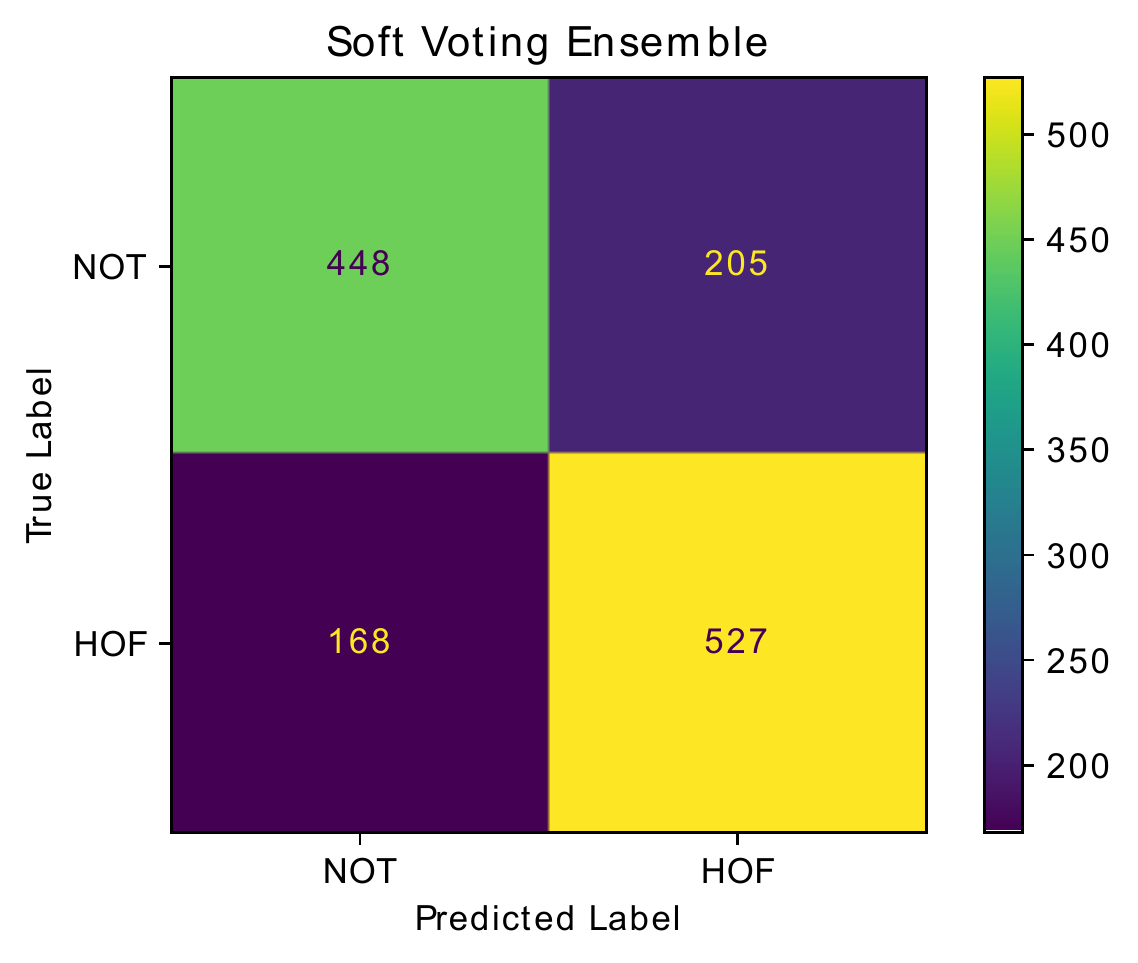}\hfill
\includegraphics[width=.18\textwidth]{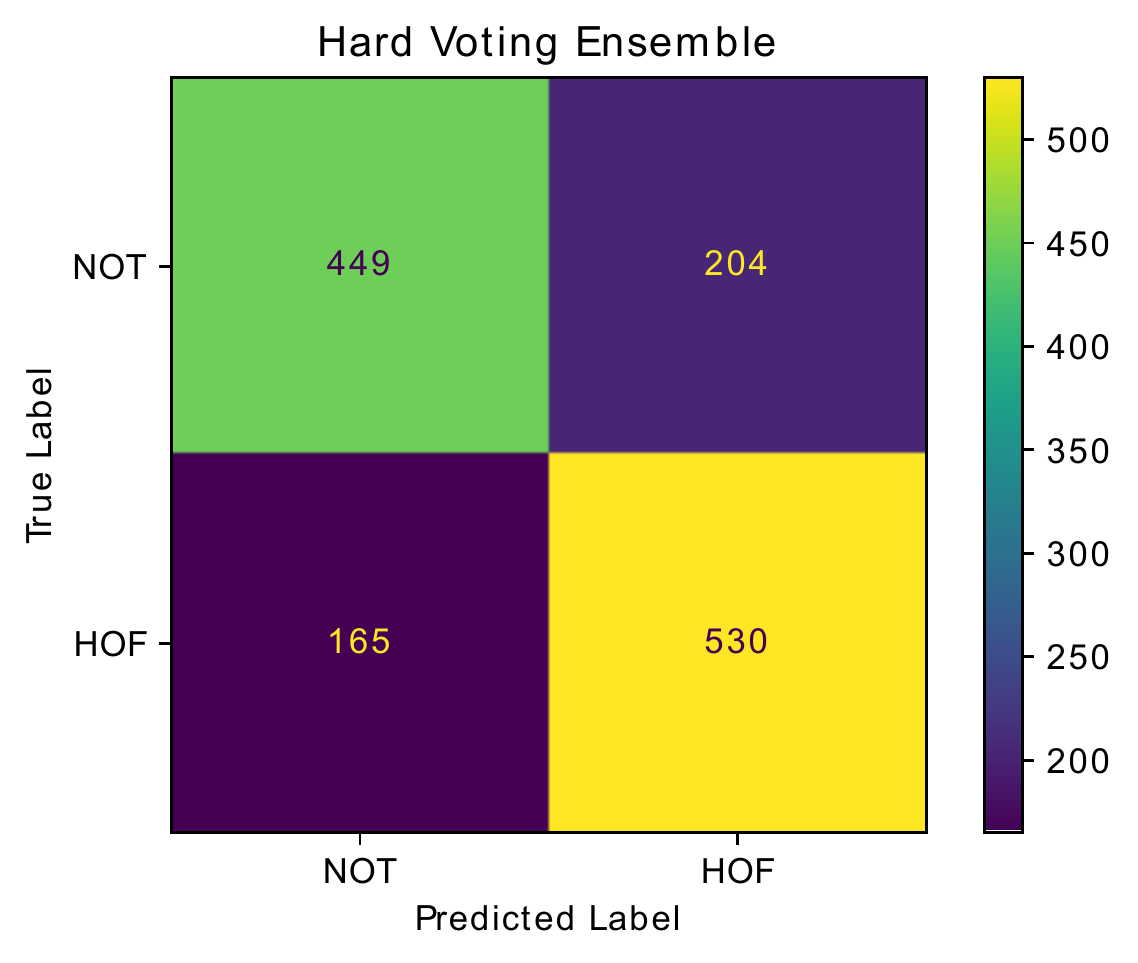}\hfill
\caption{Confusion Matrix on Test Data for Baseline and Ensemble Models}
\label{fig:cm_models}
\end{figure}



%% file: sections/conclusion.tex
\section{Conclusion}
In this paper, we deal with a novel problem of detecting hateful tweets in twitter conversations where the comment or reply might not be offensive and toxic but contributes to the hate associated with the parent tweet. We performed a thorough experimental analysis with state-of-the-art models such as XLM-RoBERTa, Indic-BERT, and Multilingual BERT to show that pre-trained multi-lingual transformer models can achieve decent performance on this task. We further demonstrate that this performance can be improved with model ensemble techniques such as Soft Voting and Hard Voting. As this problem is dealt with for the first time,  there could be many ways to improve these numbers and build a more robust system by incorporating other factors such as emojis and hashtags, which may equally or partially contribute to the hatefulness of a tweet. Additionally, we aim to explore better architectures for taking into account the context of a comment like parent comment and replies to judge the nature of the comment. Author profiling also would be a potential area of research to detect implicit hate in conversations.